\pdfoutput=1

\documentclass[11pt]{article}

\usepackage{EACL2023}

\usepackage{times}
\usepackage{latexsym}

\usepackage[T1]{fontenc}

\usepackage[utf8]{inputenc}

\usepackage{microtype}

\usepackage{inconsolata}
\newcommand\numberthis{\addtocounter{equation}{1}\tag{\theequation}}
\usepackage{amsmath, amssymb, amsfonts, amsthm}
\usepackage{xspace, graphicx, subfigure}
\newcommand{\our}{\text{CIM}\xspace}

\title{Empirical Analysis for Unsupervised Universal Dependency Parse Tree Aggregation}

\author{Adithya Kulkarni, Oliver Eulenstein \and Qi Li \\
        Department of Computer Science, Iowa State University \\
        \{aditkulk, oeulenst, qli\}@iastate.edu \\}

\begin{document}
\maketitle

\begin{abstract}

Dependency parsing is an essential task in NLP, and the quality of dependency parsers is crucial for many downstream tasks. Parsers' quality often varies depending on the domain and the language involved. Therefore, it is essential to combat the issue of varying quality to achieve stable performance. In various NLP tasks, aggregation methods are used for post-processing aggregation and have been shown to combat the issue of varying quality. However, aggregation methods for post-processing aggregation have not been sufficiently studied in dependency parsing tasks. In an extensive empirical study, we compare different unsupervised post-processing aggregation methods to identify the most suitable dependency tree structure aggregation method.
\end{abstract}

\section{Introduction}
Dependency parsers analyze the grammatical structure of a given sentence and establish the relationship between the tokens of the sentence. They are crucial for many downstream tasks, such as relation extraction~\cite{tian2021dependency} and aspect extraction~\cite{chakraborty2022open}. Dependency parsers based on large language models (LLM)~\cite{ustun-etal-2020-udapter, xu-etal-2022-multi} achieve state-of-the-art performance for several languages. However, the performance of these models on low-resource languages is limited due to their dependency on labeled data and language-specific pre-trained models. Ensemble or non-ensemble models that do not use LLMs may be suitable for low-resource languages. Therefore, it is essential to have a language and domain-agnostic framework that can estimate the quality of these input parsers and use the estimation to correct mistakes of the input parsers.

As a pioneer study, we consider the post-processing aggregation of the tree structure outputs of dependency parsers, which we call \textsl{Dependency Tree Structure (DTS)}. To improve the appeal of the aggregation in practice, we adopt the unsupervised setting, where little or no ground truth annotations are available to evaluate the qualities of the base parsers before aggregation. We compare different post-processing aggregation frameworks to address the following questions: (i) Which aggregation framework is suitable for DTS aggregation, and (ii) Can aggregation methods outperform individual state-of-the-art base parsers?

The aggregation frameworks compared in this study include MST~\cite{gavril1987generating}, Conflict Resolution on Heterogeneous Data (CRH) framework~\cite{li2014resolving}, and Customized Ising Model (\our), an extension of the classic Ising model~\cite{ravikumar2010high}. MST is a na\"ive tree aggregation method that assumes that all the parsers are of the same quality and aggregates using the maximum spanning tree (MST) algorithm. CRH framework is an optimization-based method that estimates parser quality, and \our models the joint distribution of the aggregated result and the input by estimating the quality of the input. It is designed for binary label aggregation and requires some extension for trees. For the experiments, we consider $71$ Universal Dependency (UD) test treebanks of the CONLL 2018 shared task that covers $49$ languages across different domains. As base parsers, we consider state-of-the-art ensemble, non-ensemble, and LLM-based parsers.

\section{Related Works}
The related works are summarized into the following categories:

\noindent \textbf{Label Aggregation:} These studies aggregate multiple labels obtained using labeling functions or crowd workers to quickly and cheaply annotate for large-scale unlabeled data. These works can be broadly categorized as programmatic weak supervision approaches~\cite{ratner2016data, ratner2017snorkel, ratner2019training, chen2021comparing, kuang2022firebolt, ravikumar2010high}, constraint-based weak supervision approaches~\cite{mazzetto2021adversarial, mazzetto2021semi} and optimization based approaches~\cite{li2014resolving, sabetpour2021truth}. In this work, we compare CRH framework~\cite{li2014resolving}, an optimization-based approach, and the Ising model~\cite{ravikumar2010high}, one of the weak supervision label aggregation approaches, to the aggregation of DTS.

\noindent \textbf{Tree Aggregation:} The methods in this category aggregate multiple tree structures into one representative tree. The problem of tree aggregation has been extensively studied in the phylogenetic domain~\cite{bryant2003classifica, bininda2004phylogenetic}, where trees are branching diagrams showing the evolutionary relationships among biological species. Since these mentioned methods are introduced in the phylogenetic domain, they do not consider the characteristics of parse trees. Characteristics of parse trees are considered by~\cite{kulkarni2022cptam}, where the constituency parse trees are aggregated by adopting the CRH framework~\cite{li2014resolving}. 

\noindent \textbf{Tree Ensemble:} This category contains studies that ensemble multiple input trees to obtain a representative tree. Prior studies such as random forest~\cite{probst2019hyperparameters} or boosted trees~\cite{de2007boosted} perform ensembling on the classification decisions where they depend on ground truth to learn the aggregated tree. Another line of studies~\cite{sagae2006parser, nivre2008integrating, surdeanu2010ensemble, kuncoro2016distilling} uses the maximum spanning tree to obtain aggregated trees from the weighted directed graph, where ground truth is used for weight computation.


The challenge of estimating dependency parser quality without ground truth has yet to be considered by previous studies.

\section{Comparing Aggregation Frameworks}
\label{comparing_a_f}

For aggregating parse trees, MST can be adopted. However, the aggregated results may be sub-optimal since low-quality parsers are given equal weights to high-quality parsers. Previous studies~\cite{sabetpour2021truth, kuang2022firebolt} have shown that quality estimation plays a key role in aggregating the results of multiple models. Both CRH framework and \our estimate parser quality. 

The basic idea of the CRH framework is that the inferred aggregation results are likely to be correct if supported by reliable sources. Therefore, the objective is to minimize the overall weighted distance of the aggregated results to individual sources where reliable sources have higher weights~\cite{li2014resolving}. The aggregation problem is modeled as an optimization problem and is solved by applying a block coordinate descent algorithm to estimate source quality and aggregation results iteratively. CRH framework only models the supporting votes for estimating the parser quality. Detailed discussion is provided in Appendix~\ref{app:crh_framework}.


The classic Ising model~\cite{ravikumar2010high} is a probabilistic model that aims to estimate the joint distribution of the labels provided by input sources and the unknown ground truth labels. \our also models the label correlation between the input sources; if two are correlated, their outputs are considered as a single output instead of two different outputs. This re-weighting of sources helps reduce error propagation and accurately estimate source reliability. Since it is a probabilistic model, \our considers both supporting and opposing votes to estimate source reliability. Furthermore, it does not utilize any distance measurement, and parameter estimation is solely based on the labels provided by the input sources. Detailed discussion is provided in Appendix \ref{app:ising_model}.

\section{Methodology}

Except for MST, the other two aggregation frameworks considered in this study are not designed for DTS aggregation. Both the CRH framework and \our are designed for label aggregation. Therefore, we model the DTS aggregation problem as an edge-level binary label aggregation problem, where the binary label indicates the existence of an edge. We further propose post-processing steps to ensure aggregated results follow proper DTS constraints.

\subsection{Problem Formulation}
Let $\mathcal{D} = \{s_{i}\}_{i=1}^{n}$ be a dataset with $n$ sentences. Let $T_i = \{t_{1i}, t_{2i}, \ldots, t_{qi}\}$ be the set of tokens in the sentence $s_i \in \mathcal{D}$. Let $\mathbf{P} = [P_1, P_2, P_3, \ldots, P_m]^{T}$ be $m$ dependency parsers and $\tau_{j} = \{\tau_{1j}, \tau_{2j}, \ldots, \tau_{nj}\}$ be the DTSs obtained for the dataset $\mathcal{D}$ using dependency parser $P_j$. Therefore, for a sentence $s_i$, $S_{i} = \{\tau_{i1}, \tau_{i2}, \ldots, \tau_{im}\}$\footnote{We consider that all $\tau_{ij}$'s for the sentence $s_{i}$ have the same vertex set $T_i$. Therefore, we assume all DTSs $\tau_{ij}$'s in $S_{i}$ follow the same token segmentation.} is the set of all DTSs obtained using dependency parsers $\mathbf{P}$. Each DTS, $\tau_{ij} = (T_i, E_{ij}) \in S_{i}$ is a dependency tree structure with the tokens of the sentence $s_i$ as vertices and edges connect the dependent tokens. For each $s_i \in \mathcal{D}$, DTS aggregation aims to aggregate $S_{i}$ into one representative DTS.

\subsection{Edge-level Binary Label Aggregation Problem}

To convert the DTS aggregation problem into an edge-level binary label aggregation problem, we consider the $m$ dependency parsers $\mathbf{P} = [P_1, P_2, \ldots, P_m]^{T}$ as the labeling functions\footnote{A labeling function (LF) is a mathematical function that takes $x_i$ as input and provides a label as output.} $\mathbf{L} = [L_1, L_2, \ldots, L_m]^{T}$. For each sentence $s_i \in \mathcal{D}$, the DTSs $\tau_{ij} \in S_{i}$ differ only concerning the edges. We utilize this observation to define the DTS aggregation problem as an edge-level binary label aggregation problem. Specifically, let $\mathcal{E}_{i} = \cup_{j=1}^{m} E_{ij}$ be the union of edge sets $E_{ij}$ from each DTS $\tau_{ij} \in S_{i}$ and $\mathbf{E} = \cup_{i=1}^{n}\mathcal{E}_{i}$ be the union of all $\mathcal{E}_{i}$ for the dataset $\mathcal{D}$. For aggregation, we consider each $e \in \mathbf{E}$ as an instance of $\mathcal{D}$. On the sentence level, the binary labels for each $e \in \mathcal{E}_{i}$ using the LF $L_j \in \mathbf{L}$ is obtained as follows:
\begin{align*}
  L_j(e) =\begin{cases}
    1, & \text{if $e \in E_{ij}$}\\
    -1, & \text{$Otherwise$} \numberthis
  \end{cases}.
  \label{eq1}
\end{align*}
Thus, the DTS aggregation problem is converted into a binary labeling task, and both the CRH framework and \our can be applied to aggregate labels. Since \our considers label correlation between input parsers, below we discuss how \our is used for DTS aggregation.

\subsection{\our for Dependency Tree Structure Aggregation}

\our models the label correlation between the input parsers. Due to the absence of ground truth labels, we estimate the label correlation using the majority voting results of the labels from the input parsers. Then, the label correlation between the input parsers is estimated using an $\mathrm{l}_1$-regularized logistic regression, in which the pairwise correlation between input parsers is estimated by performing logistic regression subject to an $\mathrm{l}_1$-constraint. For more details, refer to \cite{ravikumar2010high}.


Let $\mathbf{Y}$ be the random variable denoting the unknown ground truth labels for $\mathcal{D}$. With the estimated label correlation between input parsers, the joint distribution $P_{\mu}(\mathbf{Y}, \mathbf{L})$ is estimated by learning the mean $\mu = \{\mu_{00}, \mu_{+}, \mu_{0+}, \mu_{++}\}$ and canonical $\theta = \{\theta_{00}, \theta_{+}, \theta_{0+}, \theta_{++}\}$ parameters of the model. Once the distribution $P_{\mu}(\mathbf{Y}, \mathbf{L})$ is learned, the probabilistic scores for each $x_i \in \mathcal{D}$ are inferred as:

\resizebox{.9\linewidth}{!}{
\begin{minipage}{\linewidth}
\begin{align*}
    P(\hat{y}_i=1 | \mathbf{L}(x_i); \hat{\theta}_{00}, \hat{\theta}_{0+}) = \sigma (2\hat{\theta}_{00} + 2\hat{\theta}_{0+}\mathbf{L}(x_i)), \numberthis
    \label{eq3}
\end{align*}
\end{minipage}
}
where $\mathbf{L}(x_i)$ is the $i-th$ row in $\mathbf{L}$ representing the set of $m$ labels obtained for $x_i$ using LFs, $\hat{y}_{i}$ is the aggregated label for $x_i$, and $\sigma(z) = \frac{1}{1+\exp(-z)}$. The probabilistic scores encompass parser quality. Then, for each $S_{i}$, we obtain a weighted token graph ${\omega}_{i} = (T_i, \mathcal{E}_{i})$. We update the edge weights of the token graph $\omega_i$ with the inferred probability scores and apply the MST on the updated $\omega_i$ to ensure the final aggregation results follow tree structure constraints. 

\section{Experiments}

For the experiments, we aim to have sufficient diversity concerning the treebanks and the base parsers. Therefore, we empirically test MST, CRH framework, and \our on $71$ UD test treebanks of CONLL 2018 shared task~\cite{zeman2018conll}. These treebanks cover $49$ languages across different domains. The shared task also provides outputs of the participating teams on these test treebanks\footnote{The outputs of the participating dependency parsers on the test treebanks of the CoNLL 2018 shared task is archived and made public at \url{http://hdl.handle.net/11234/1-2885}.}. The participating teams include various ensemble and non-ensemble methods. We directly utilize these outputs as parser predictions. In addition to these methods, we consider two state-of-the-art LLM-based methods~\cite{ustun-etal-2020-udapter, xu-etal-2022-multi}. We re-train these models on the train set of the shared task and obtain outputs for the test treebanks. We pre-process the outputs to ensure all parser outputs have the same token segmentation; the detailed steps are provided in Appendix \ref{app:data_preprocessing}. We highlight the summary of the results in this section\footnote{Please refer to Section \ref{app:full_results} and Section \ref{app:comparitive_study} in Appendix~\ref{sec:appendix} for full results and comparison study, respectively.}.

\subsection{Experimental Setup}

In real-life scenarios, it is common practice to estimate the quality of dependency parsers by evaluating their outputs on small annotated samples and then using high-quality parsers to save computation costs. To align with the real-life applications, for each of the $71$ pre-processed test treebanks, we sample $10$ sentences and rank all the input parsers, including ensemble methods, depending on the performance of their output DTSs on these sentences. Then, we choose the top 9 dependency parsers to include diverse methods (ensemble, non-ensemble, and LLM-based) in the aggregation.

\subsection{Evaluation Metrics}
We use the Unlabeled Attachment Score (UAS) for evaluation, which considers the percentage of nodes with correctly assigned references to the parent node. We compute the Mean ($\mu$), Median (M), and standard deviation ($\sigma$) of UAS scores on $71$ test treebanks.

\subsection{Baseline Methods}

The three aggregation frameworks are compared with top two ensemble methods, including HIT-SCIR~\cite{che2018towards} and LATTICE~\cite{lim2018sex}, top two non-ensemble methods, including TurkuNLP~\cite{kanerva2018turku} and UDPipe Future~\cite{straka2018udpipe} from CoNLL 2018 shared task, and two LLM-based methods, UDapter~\cite{ustun-etal-2020-udapter} and MLPSBM~\cite{xu-etal-2022-multi}. Following their respective papers' settings, we use the \textit{BERT-multilingual-cased} encoder for both UDapter and MLPSBM. MLPSBM is tested only for high-resource languages such as Bulgarian, Catalan, Czech, German, English, Spanish, Italian, Dutch, Norwegian, Romanian, and Russian. For reference, we also present the performance of the best parser among these top 9 chosen parsers (BEST) and the average performance of the top 9 chosen parsers (Average) for each test treebank evaluated using ground truth annotations.

\subsection{Results and Discussion}

Table~\ref{table: main_table} and \ref{table: main_table_1} compare three aggregation frameworks with the baselines for high-resource and low-resource language treebanks, respectively. Comparing the aggregation methods and ensemble methods, we can observe that aggregation methods, in general, achieve better performance than ensemble methods, with higher mean UAS scores and lower standard deviation across different test treebanks. The better performance is thanks to the flexibility for aggregation methods to choose different base models. Similar observations can be made when comparing aggregation methods with non-ensemble methods. Comparing the aggregation methods with LLM-based methods, we can observe that only \our outperforms the LLM-based methods, suggesting that it is the most suitable DTS aggregation framework. Using the updated UAS score as distance measurement for the CRH model may be a sub-optimal choice for DTS aggregation since it only considers the supporting votes for each edge. Thus, the CRH model cannot estimate the parser quality properly and perform similarly to MST, which does not consider parser quality. We can observe that \our can outperform all baseline methods in terms of mean and median in both Tables~\ref{table: main_table} and \ref{table: main_table_1} and standard deviation in Table~\ref{table: main_table_1}, even comparing with the best parser among the chosen top 9 for each test treebank.

\begin{table}[t]
\caption{Results of MST, CRH, and \our compared with the baselines for high-resource language treebanks. The best performance is highlighted in \textbf{bold}, and the runner-up is highlighted with \underline{underline}.}
\small
\centering
\begin{tabular}{c|c|c|c}
\hline
\textbf{Method} & \textbf{$\mu $(UAS)} & \textbf{M (UAS)} & \textbf{$\sigma$ (UAS)} \\
\hline
HIT-SCIR & 87.37 & 88.83 & 5.27 \\
LATTICE & 83.01 & 87.84 & 15.10 \\ \hline
TurkuNLP & 80.95 & 86.43 & 15.77 \\ 
UDPipe Future & 80.21 & 84.99 & 14.04 \\ \hline
UDapter & 89.43 & 90.1 & 4.45 \\ 
MLPSBM & \underline{93.04} & \underline{93.92} & \textbf{3.13} \\ \hline
Average & 82.5 & 85.11 & 8.74 \\
BEST & \underline{93.04} & \underline{93.92} & \textbf{3.13} \\ \hline
MST & 88.42 & 90.12 & 5.37 \\ 
CRH & 87.23 & 88.97 & 4.91 \\
\our & \textbf{93.18} & \textbf{94.02} & \underline{3.2} \\
\hline
\end{tabular}
\label{table: main_table}
\end{table}

\begin{table}[t]
\caption{Results of MST, CRH, and \our compared with the baselines for low-resource language treebanks. The best performance is highlighted in \textbf{bold}, and the runner-up is highlighted with \underline{underline}.}
\small
\centering
\begin{tabular}{c|c|c|c}
\hline
\textbf{Method} & \textbf{$\mu $(UAS)} & \textbf{M (UAS)} & \textbf{$\sigma$ (UAS)} \\
\hline
HIT-SCIR & 78.15 & 85.64 & 17.71 \\
LATTICE & 74.55 & 82.93 & 16.64 \\ \hline
TurkuNLP & 73.48 & 82.42 & 17.96 \\ 
UDPipe Future & 74.59 & 81.63 & 16.28 \\ \hline
UDapter & 83.14 & 86.25 & 11.43 \\ \hline
Average & 74.45 & 81.19 & 14.79 \\
BEST & \underline{84.08} & \underline{87.78} & 10.13 \\ \hline
MST & 81.71 & 85.16 & 10.84 \\ 
CRH & 81.39 & 85.14 & \underline{9.98} \\
\our & \textbf{85.93} & \textbf{89.33} & \textbf{9.68} \\
\hline
\end{tabular}
\vspace{-0.5cm}
\label{table: main_table_1}
\end{table}

\section{Conclusion}

This empirical study compares three aggregation frameworks, MST, CRH, and \our, for the task of DTS aggregation. We model the DTS aggregation problem as an edge-level binary label aggregation problem to employ CRH and \our, which are specifically designed for label aggregation. Extensive empirical studies on $71$ UD test treebanks of CONLL 2018 shared task demonstrate that \our is the most suitable DTS aggregation method that can properly estimate parser quality and outperform state-of-the-art base parsers. We will consider the aggregation of relation labels in future work.

\section{Limitations}
In this work, we only consider the tree structure of the dependency parse tree under the assumption that each sentence has the same token segmentation across the input dependency parse trees. If the token segmentation differs across the input dependency parse trees, our proposed approach is not applicable. Furthermore, our proposed approach is not tested to aggregate different relation labels of dependency parse trees. 

\section{Ethics Statement}

We comply with the ACL Code of Ethics.

\bibliography{custom}
\bibliographystyle{acl_natbib}

\appendix

\section{Appendix}
\label{sec:appendix}

\subsection{CRH Framework}
\label{app:crh_framework}

The CRH framework is an optimization framework to minimize the overall distance of the aggregated result to a reliable source~\cite{li2014resolving}. The optimization framework is defined as:
\begin{align*}
\min_{\mathcal{Y}^*,\mathcal{W}} f(\mathcal{Y}^*,\mathcal{W})=\sum_{k=1}^{m} {w_k} \sum_{i=1}^{n} \sum_{j=1}^{q} d_j(l_{ij}^*,l_{ij}^k)\\
s.t. \ {\delta(\mathcal W)=1} \numberthis
\label{optimization}
\end{align*}

where $\mathcal{Y}^*$ and $\mathcal{W}$ represent the set of aggregated truths and the source weight, respectively, $w_k$ refers to the reliability degree of the $k$-th source. The distance measurement function $d_j(\cdot,\cdot)$ measures the distance between the labels provided by the sources $l_{ij}^k$ and the aggregated labels $l_{ij}^*$. The regularization function $\delta(\mathcal{W})$ is defined to ensure that the weights are always non-zero and positive.

The block coordinate descent algorithm is applied to optimize the objective function in Eq.~(\ref{optimization}) by iteratively updating the source weights and aggregated truths by following the two steps below.

\textbf{Step 1: Source Weight Update.}

The source weights are updated considering the values for the aggregated truths as fixed. The updated source weights are computed following Eq.~(\ref{Weight Update}) that jointly minimize the objective function.

\resizebox{.90\linewidth}{!}{
\begin{minipage}{\linewidth}
\begin{align*}
     \mathcal{W} \leftarrow \operatorname*{argmin}_\mathcal{W} f(\mathcal{Y}^*,\mathcal{W})\  s.t. \  {\delta(\mathcal W)=\sum_{k=1}^{m}exp(-w_k)}. \numberthis
\label{Weight Update}
\end{align*}
\end{minipage}
}

Eq.~(\ref{Weight Update}) regularizes the value of $w_k$ by constraining the sum of $exp({-w_k})$.

\textbf{Step 2: Aggregated Truth Update.}

To update the aggregated truths, the weight of each source $w_k$ is considered fixed. The aggregated truths are updated following Eq.~(\ref{Truth Update}) that minimizes the difference between the truth and the sources' labels, where sources are weighted by their estimated reliabilities.

\begin{align*}
{l_{im}^{(*)}} \leftarrow
\operatorname*{argmin}_{l} 
{\sum_{k=1}^{m}w_k\cdot d_m(l,l_{ij}^k)}. \numberthis
\label{Truth Update}
\end{align*}

Eq.~(\ref{Truth Update}) provides the collection of aggregated truths $\mathcal{Y}^*$ that minimize $f(\mathcal{Y}^*,\mathcal{W})$ with fixed $\mathcal{W}$.

\subsection{Ising Model}
\label{app:ising_model}

The Ising model~\cite{ravikumar2010high} is proposed to obtain aggregated labels for a binary labeling task. Let $\mathcal{D} = \{x_{i}\}_{i=1}^{n}$ be a dataset with $n$ instances. Let $\mathbf{L} = [L_1, L_2, ..., L_m]^{T}$ be an $n \times m$ matrix containing the labels provided by $m$ binary labeling functions (LFs) for the instances in dataset $\mathcal{D}$. Let the unobserved ground truth label for each $x_i$ be $y_i \in \{-1, 1\}$. We use $\mathbf{Y}$ to denote the random variable for the ground truth labels for the dataset $\mathcal{D}$. The Ising model aims to estimate the joint distribution $P_{\mu}(\mathbf{Y}, \mathbf{L})$ and learn the parameters $\mu$. 

To consider the correlations between the LFs, the Ising model can take an undirected correlation graph $G = (V, E)$ as an additional input. In this correlation graph, vertices are the LFs and the ground truth random variable $\mathbf{Y}$, $V = \{L_1, L_2, L_3, ..., L_m, \mathbf{Y}\}$, and an edge $e \in E$ indicates that the connected vertices are correlated.
Each LF has an edge to $\mathbf{Y}$ since each LF contributes to estimating the random variable $\mathbf{Y}$ and is thus correlated to $\mathbf{Y}$. With this correlation graph $G$, the joint distribution between $\mathbf{Y}$ and LFs $\mathbf{L}$ is estimated by the Ising model as:
\resizebox{.90\linewidth}{!}{
\begin{minipage}{\linewidth}
\begin{align*}
    P_{\mu}(\mathbf{Y}, \mathbf{L}) = \frac{1}{Z} \exp(&\theta_{00}\mathbf{Y} + \sum_{j=1}^{m}\theta_{jj}L_{j} + \sum_{j=1}^{m}\theta_{0j}L_{j}\mathbf{Y} \\ +& \sum_{(L_j, L_k)=1}\theta_{jk}L_{j}L_{k}),\numberthis
    \label{app:eq1}
\end{align*}
\end{minipage}
}
\noindent where $Z$ is a partition function ensuring that the distribution sums to one and $\theta = \{\theta_{00}, \theta_{+}, \theta_{0+}, \theta_{++}\}$ are the canonical parameters. For each canonical parameter there is an associated mean parameter $\mu = \{\mu_{00}, \mu_{+}, \mu_{0+}, \mu_{++}\}$. Together, the canonical and mean parameters reflect the quality of the LFs. To compute $P_{\mu}(\mathbf{Y}, \mathbf{L})$, the mean parameters are learned first, and then they are used to learn the canonical parameters by solving the following logistic regression problem:
\begin{align*}
    \hat{\theta}_{00}, \hat{\theta}_{0+} =& \arg \min_{\theta_{00}, \theta_{0+}} -\theta_{00}\mu_{00} - \theta_{0+}^{T}\mu_{0+} \\ 
    &+ \frac{1}{n}\sum_{i=1}^{n} 
    \log [\exp(\theta_{00} + \theta_{0+}^{T}\mathbf{L}(x_i)) \\
    &+ \exp(-\theta_{00} - \theta_{0+}^{T}\mathbf{L}(x_i))]. \numberthis
    \label{app:eq2}
\end{align*}
Once the distribution $P_{\mu}(\mathbf{Y}, \mathbf{L})$ is learned, the probabilistic scores for each $x_i \in \mathcal{D}$ are inferred as:

\resizebox{.90\linewidth}{!}{
\begin{minipage}{\linewidth}
\begin{align*}
    P(\hat{y}_i=1 | \mathbf{L}(x_i); \hat{\theta}_{00}, \hat{\theta}_{0+}) = \sigma (2\hat{\theta}_{00} + 2\hat{\theta}_{0+}\mathbf{L}(x_i)), \numberthis
    \label{app:eq3}
\end{align*}
\end{minipage}
}
where $\mathbf{L}(x_i)$ is the i-th row in $\mathbf{L}$ representing the set of $m$ labels obtained for $x_i$ using LFs, $\hat{y}_{i}$ is the aggregated label for $x_i$, and $\sigma(z) = \frac{1}{1+\exp(-z)}$. For further details, refer to~\citet{kuang2022firebolt}.

\subsection{Data Pre-processing}
\label{app:data_preprocessing}
We obtain the existing outputs of the $26$ participating teams on $82$ UD test treebanks from the CoNLL 2018 shared task organizers, plus the outputs of two LLM-based parsers. We discard sentences from the test treebanks where input parsers provide different token segmentation. To better illustrate the performance of \our, sentences with total agreement on the DTS from all the input dependency parsers are regarded as easy sentences and also discarded. To obtain a statistically meaningful evaluation, we discard treebanks with less than $50$ sentences or less than $9$ participating parsers. Eventually, we obtain $71$ UD test treebanks used in our experiments\footnote{Refer to Table \ref{tab:full_results1}, \ref{tab:full_results2} and \ref{tab:full_results3} for the test treebank details.}.

\subsection{Full Results}
\label{app:full_results}
Full results\footnote{We will publish the code upon acceptance} for $19$ high-resource language treebanks are shown in Table \ref{tab:full_results1} and the full results for $52$ low-resource language treebanks are shown in Tables \ref{tab:full_results2} and \ref{tab:full_results3}. The table is ordered alphabetically concerning the name of the treebank. From the results in Table \ref{tab:full_results1}, taking treebank $bg\_btb$ as an example, we can observe that MLPSBM, the LLM-based parser, achieved the best results among the baselines and \our outperformed all the baselines for this treebank. Similar observation can be made for results in Table \ref{tab:full_results2} for $af\_afribooms$ treebank where UDapter, another LLM-based parser, achieved the best results among the baselines and \our outperformed all the baselines for this treebank too. The examples confirm that \our is the most suitable DTS aggregation framework that can properly estimate parser quality and outperform state-of-the-art baselines.

\subsection{Comparison Study}
\label{app:comparitive_study}
\begin{figure}
    \centering
    \includegraphics[width=0.49\textwidth]{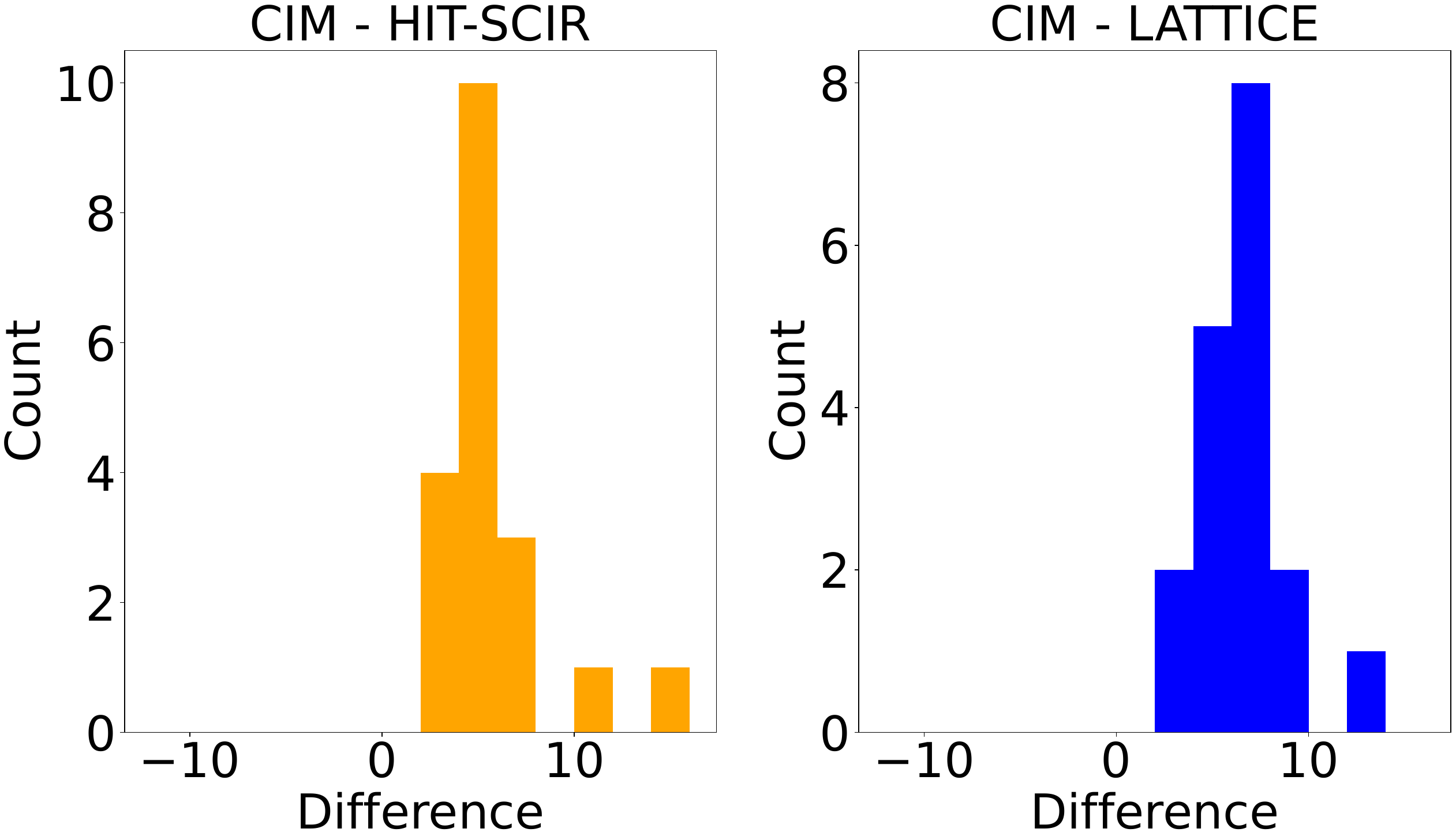}
    \caption{Difference between \our and ensemble baselines}
    \label{fig:fig1}
\end{figure}

\begin{figure}
    \centering
    \includegraphics[width=0.49\textwidth]{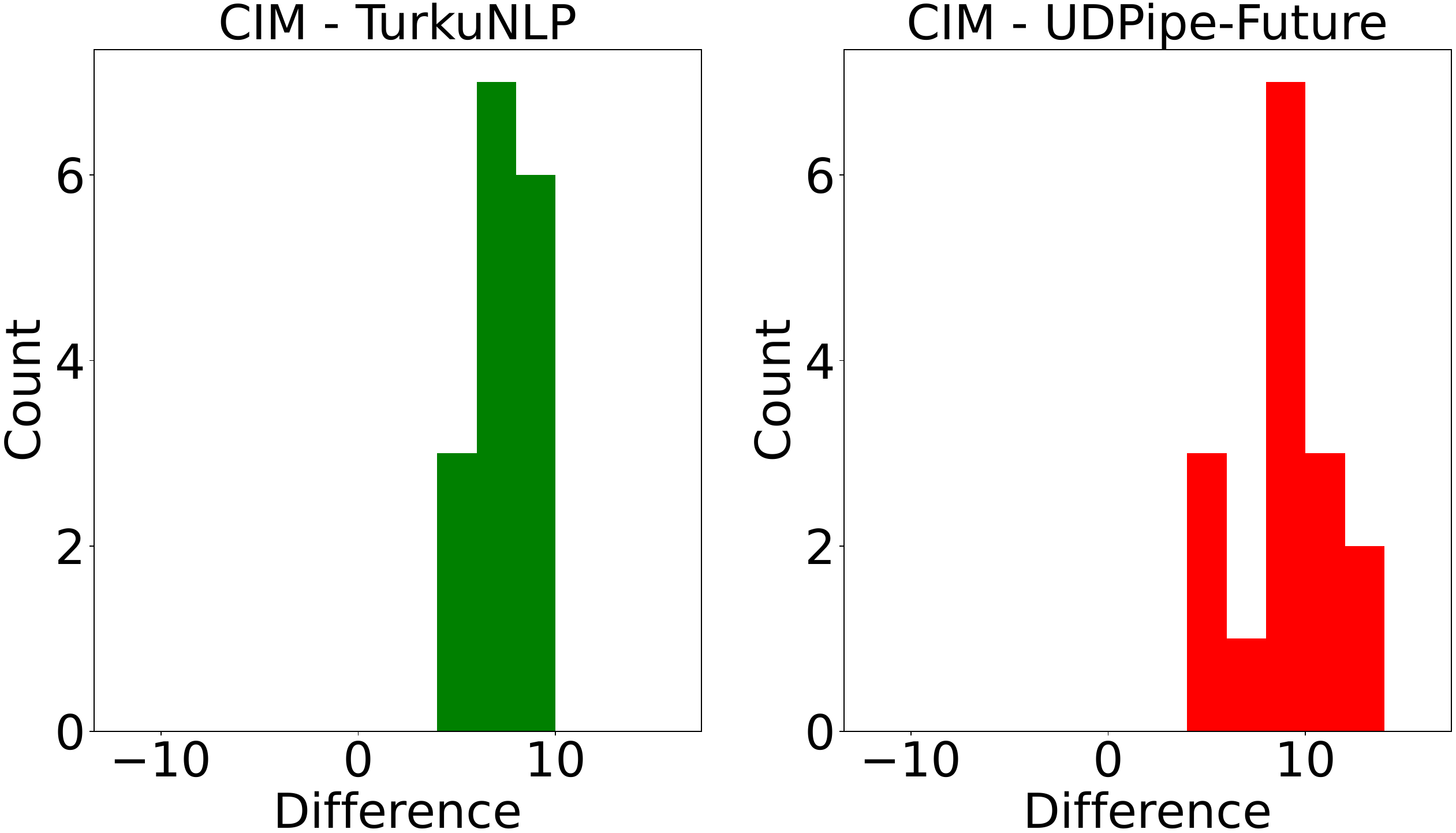}
    \caption{Difference between \our and non-ensemble baselines}
    \label{fig:fig2}
\end{figure}

\begin{figure}
    \centering
    \includegraphics[width=0.49\textwidth]{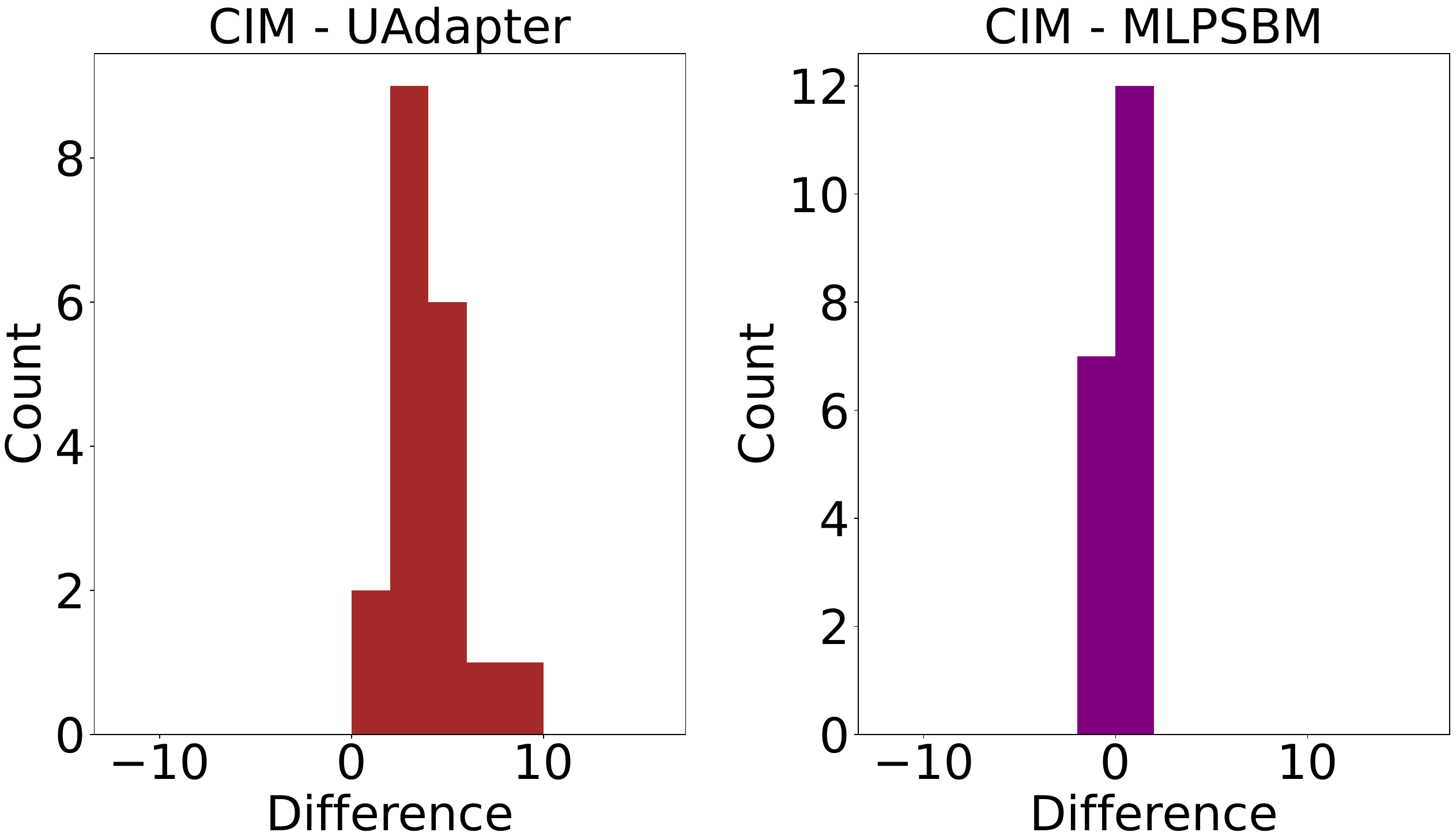}
    \caption{Difference between \our and LLM-based baselines}
    \label{fig:fig3}
\end{figure}

Figures~\ref{fig:fig1}, \ref{fig:fig2}, and \ref{fig:fig3} depict the histograms of comparisons between \our with ensemble, non-ensemble, and LLM-based baselines for high-resource language treebanks. A positive difference means that \our outperforms. From figure \ref{fig:fig1}, we can observe that \our outperforms ensemble methods, including HIT-SCIR and LATTICE. From figure \ref{fig:fig2}, we can observe that \our outperforms non-ensemble methods, including TurkuNLP and UDPipe Future. From figure \ref{fig:fig3}, we can observe that \our outperforms LLM-based baselines UDapter and MLPSBM. These results support our claim that \our can correct the mistakes of individual parsers --- even the best ones --- and outperform them. These results support our claim that \our is the most suitable DTS aggregation framework among the compared aggregation frameworks in this study.

\begin{table*}[t]
\caption{Full results of MST, CRH, \our compared with the baselines for $19$ high-resource language treebanks. The table is ordered in alphabetical order with respect to the name of the treebank. The columns E1 and E2 refer to ensemble baselines HIT-SCIR~\cite{che2018towards}, and LATTICE~\cite{lim2018sex}, respectively. The columns NE1, NE2 refer to non-ensemble baselines TurkuNLP~\cite{kanerva2018turku} and UDPipe Future~\cite{straka2018udpipe}, respectively. The columns L1 and L2 refer to LLM-based parsers UDapter~\cite{ustun-etal-2020-udapter} and MLPSBM~\cite{xu-etal-2022-multi}, respectively. The column Avg. refers to the baseline Average.}
\centering
\begin{tabular}{c|c|c|c|c|c|c|c|c|p{0.74cm}|p{0.64cm}|p{0.64cm}|p{0.64cm}}
\hline
\textbf{TB} & \textbf{Sent.} & \textbf{E1} & \textbf{E2} & \textbf{NE1} & \textbf{NE2} & \textbf{L1} & \textbf{L2} & \textbf{Avg.} & \textbf{BEST} & \textbf{MST} & \textbf{CRH} & \textbf{\our} \\ \hline
bg\_btb            & 226                & 91.4          & 88.3          & 89.2            & 87.1            & 90.1              & 95.7            & 89.6          & 95.7          & 92.1         & 91.0         & 96.1          \\ \hline
ca\_ancora         & 364                & 91.4          & 90.0          & 89.6            & 90.0            & 92.6              & 95.7            & 83.3          & 95.7          & 92.5         & 91.2         & 94.6          \\ \hline
cs\_cac            & 156                & 88.5          & 87.8          & 88.1            & 87.3            & 89.8              & 95.2            & 88.8          & 95.2          & 90.1         & 88.9         & 95.          \\ \hline
de\_gsd            & 212                & 83.6          & 81.2          & 81.1            & 78.3            & 86.2              & 89.3            & 82.2          & 89.3          & 83.3         & 80.2         & 89.9          \\ \hline
en\_ewt            & 340                & 89.3          & 88.6          & 87.1            & 81.1            & 90.2              & 93.4            & 86.3          & 93.4          & 88.6         & 87.3         & 93.6          \\ \hline
en\_gum            & 199                & 86.9          & 90.8          & 84.1            & 85.0            & 91.3              & 92.8            & 80.6          & 92.8          & 88.4         & 86.9         & 93.4          \\ \hline
en\_lines          & 286                & 84.6          & 85.1          & 82.6            & 79.0            & 87.2              & 89.4            & 83.3          & 89.4          & 83.3         & 81.9         & 90.4          \\ \hline
en\_pud            & 264                & 86.7          & 88.7          & 84.0            & 80.5            & 88.4              & 91.3            & 84.9          & 91.3          & 87.2         & 86.9         & 92.5          \\ \hline
es\_ancora         & 387                & 89.8          & 88.7          & 88.2            & 88.3            & 88.9              & 93.9            & 88.9          & 93.9          & 91.3         & 89.5         & 93.6          \\ \hline
it\_isdt           & 83                 & 92.7          & 91.3          & 90.3            & 90.9            & 93.9              & 96.4            & 85.1          & 96.4          & 93.1         & 92.6         & 95.5          \\ \hline
it\_postwita       & 136                & 91.3          & 21.8          & 20.5            & 27.2            & 92.7              & 95.2            & 57.5          & 95.2          & 92.6         & 86.6         & 95.0          \\ \hline
nl\_alpino         & 152                & 88.6          & 86.5          & 85.4            & 85.6            & 89.2              & 92.9            & 87.2          & 92.9          & 89.3         & 89.5         & 93.6          \\ \hline
nl\_lassysmall     & 116                & 88.2          & 86.2          & 84.5            & 85.0            & 89.8              & 93.9            & 80.0          & 93.9          & 89.7         & 89.0         & 94.1          \\ \hline
no\_bokmaal        & 466                & 90.6          & 88.2          & 87.7            & 86.7            & 91.2              & 94.5            & 89.1          & 94.5          & 91.3         & 90.1         & 94.9          \\ \hline
no\_nynorsk        & 378                & 90.1          & 87.3          & 86.4            & 84.7            & 91.8              & 94.6            & 88.3          & 94.6          & 90.9         & 90.3         & 95.2          \\ \hline
no\_nynorsklia     & 374                & 70.6          & 71.7          & 61.3            & 62.9            & 72.4              & 82.6            & 64.2          & 82.6          & 70.5         & 72.6         & 81.3          \\ \hline
ro\_rrt            & 199                & 88.8          & 87.3          & 87.7            & 85.9            & 89.2              & 93.2            & 88.3          & 93.2          & 91.7         & 90.4         & 94.0          \\ \hline
ru\_syntagrus      & 1150               & 89.9          & 88.0          & 88.5            & 86.0            & 92.4              & 95.3            & 89.3          & 95.3          & 92.7         & 91.8         & 94.6          \\ \hline
ru\_taiga          & 295                & 77.0          & 79.7          & 71.7            & 72.7            & 91.9              & 92.6            & 70.5          & 92.6          & 81.5         & 80.7         & 92.8          \\ \hline
\end{tabular}
\label{tab:full_results1}
\end{table*}

\begin{table*}[t]
\caption{Full results of MST, CRH, \our compared with the baselines for $52$ low-resource language treebanks. The table is ordered in alphabetical order with respect to the name of the treebank. The columns E1 and E2 refer to ensemble baselines HIT-SCIR~\cite{che2018towards}, and LATTICE~\cite{lim2018sex}, respectively. The columns NE1, NE2 refer to non-ensemble baselines TurkuNLP~\cite{kanerva2018turku} and UDPipe Future~\cite{straka2018udpipe}, respectively. The column L1 refers to LLM-based parser UDapter~\cite{ustun-etal-2020-udapter}. The column Avg. refers to the baseline Average.}
\centering
\begin{tabular}{c|c|c|c|c|c|c|c|c|c|c|c}
\hline
\textbf{TB}   & \textbf{Sent.} & \textbf{E1} & \textbf{E2} & \textbf{NE1} & \textbf{NE2} & \textbf{L1} & \textbf{Avg.} & \textbf{BEST} & \textbf{MST} & \textbf{CRH} & \textbf{\our} \\ \hline
af\_afribooms & 158            & 86.0        & 86.1        & 85.9         & 86.0         & 87.2        & 77.4          & 87.2          & 87.2         & 86.3         & 90.9                         \\ \hline
ar\_padt      & 55             & 70.1        & 70.4        & 68.8         & 68.9         & 85.6        & 73.3          & 88.8          & 90.1         & 88.4         & 92.1                         \\ \hline
br\_keb       & 249            & 17.4        & 49.8        & 48.8         & 38.5         & 60.8        & 43.1          & 65.6          & 47.7         & 64.7         & 70.2                         \\ \hline
bxr\_bdt      & 763            & 35.2        & 40.4        & 31.7         & 32.8         & 35.8        & 36.1          & 46.8          & 48.4         & 44.5         & 51.4                         \\ \hline
cs\_fictree   & 235            & 89.3        & 88.9        & 88.3         & 86.5         & 90.2        & 87.8          & 90.2          & 88.6         & 89.4         & 92.2                         \\ \hline
cs\_pdt       & 1708           & 90.2        & 88.5        & 87.6         & 85.5         & 89.4        & 87.0          & 90.2          & 88.3         & 88.1         & 91.4                         \\ \hline
cs\_pud       & 166            & 88.4        & 86.4        & 85.0         & 85.7         & 89.1        & 85.8          & 89.1          & 89.2         & 87.5         & 91.7                         \\ \hline
cu\_proiel    & 109            & 74.2        & 46.1        & 44.3         & 87.9         & 82.4        & 61.0          & 87.9          & 87.4         & 85.2         & 92.6                         \\ \hline
da\_ddt       & 195            & 86.4        & 77.7        & 77.4         & 81.5         & 88.5        & 81.0          & 88.5          & 85.1         & 83.5         & 88.9                         \\ \hline
el\_gdt       & 131            & 90.1        & 86.8        & 88.1         & 89.2         & 90.6        & 87.3          & 90.7          & 90.5         & 90.5         & 92.4                         \\ \hline
et\_edt       & 803            & 85.6        & 83.7        & 83.9         & 81.9         & 86.3        & 83.8          & 86.3          & 84.8         & 83.6         & 88.2                         \\ \hline
eu\_bdt       & 688            & 82.9        & 81.2        & 81.9         & 82.0         & 83.8        & 81.8          & 83.8          & 82.3         & 81.2         & 86.2                         \\ \hline
fa\_seraji    & 204            & 89.8        & 88.0        & 87.6         & 87.6         & 89.1        & 86.3          & 89.8          & 88.2         & 88.4         & 92.3                         \\ \hline
fi\_ftb       & 455            & 85.7        & 86.1        & 86.0         & 80.8         & 90.4        & 83.6          & 90.4          & 84.9         & 86.6         & 90.9                         \\ \hline
fi\_pud       & 203            & 87.6        & 86.1        & 83.6         & 81.6         & 91.3        & 85.4          & 91.3          & 86.8         & 87.1         & 91.6                         \\ \hline
fi\_tdt       & 386            & 84.9        & 80.0        & 78.9         & 80.6         & 88.5        & 81.2          & 88.5          & 85.0         & 85.6         & 90.2                         \\ \hline
fo\_oft       & 859            & 51.3        & 47.1        & 37.3         & 51.6         & 69.6        & 53.3          & 69.6          & 59.1         & 67.0         & 69.4                         \\ \hline
fr\_gsd       & 89             & 88.9        & 88.6        & 87.9         & 85.9         & 90.6        & 85.1          & 90.6          & 89.9         & 88.6         & 91.4                         \\ \hline
fr\_sequoia   & 51             & 88.9        & 91.0        & 87.3         & 82.7         & 92.1        & 78.5          & 92.1          & 87.8         & 88.5         & 91.7                         \\ \hline
fro\_srcmf    & 362            & 79.9        & 82.9        & 79.9         & 82.6         & 80.3        & 81.2          & 82.9          & 84.0         & 83.8         & 87.2                         \\ \hline
gl\_treegal   & 143            & 82.8        & 82.9        & 82.7         & 83.2         & 82.3        & 82.4          & 83.2          & 82.8         & 83.0         & 88.7                         \\ \hline
grc\_perseus  & 544            & 80.6        & 74.4        & 74.2         & 72.7         & 82.5        & 74.9          & 82.5          & 75.5         & 75.8         & 82.7                         \\ \hline
grc\_proiel   & 110            & 81.6        & 66.6        & 66.4         & 61.2         & 81.7        & 64.0          & 81.7          & 80.3         & 78.2         & 86.1                         \\ \hline
he\_htb       & 55             & 34.5        & 32.0        & 33.0         & 32.8         & 88.6        & 49.4          & 91.5          & 91.6         & 88.1         & 94.2                         \\ \hline
hi\_hdtb      & 346            & 92.0        & 90.7        & 91.0         & 90.1         & 92.8        & 91.0          & 92.8          & 91.7         & 90.8         & 93.5                         \\ \hline
hr\_set       & 304            & 88.6        & 87.5        & 88.1         & 83.5         & 89.4        & 87.3          & 89.4          & 88.7         & 88.0         & 91.8                         \\ \hline
\end{tabular}
\label{tab:full_results2}
\end{table*}

\begin{table*}[t]
\caption{Full results of MST, CRH, \our compared with the baselines for $52$ low-resource language treebanks (Contd.). The table is ordered in alphabetical order with respect to the name of the treebank. The columns E1 and E2 refer to ensemble baselines HIT-SCIR~\cite{che2018towards}, and LATTICE~\cite{lim2018sex}, respectively. The columns NE1, NE2 refer to non-ensemble baselines TurkuNLP~\cite{kanerva2018turku} and UDPipe Future~\cite{straka2018udpipe}, respectively. The column L1 refers to LLM-based parser UDapter~\cite{ustun-etal-2020-udapter}. The column Avg. refers to the baseline Average.}
\centering
\begin{tabular}{c|c|c|c|c|c|c|c|c|c|c|c}
\hline
\textbf{TB}   & \textbf{Sent.} & \textbf{E1} & \textbf{E2} & \textbf{NE1} & \textbf{NE2} & \textbf{L1} & \textbf{Avg.} & \textbf{BEST} & \textbf{MST} & \textbf{CRH} & \textbf{\our} \\ \hline
hsb\_ufal     & 391            & 55.0        & 60.7        & 44.8         & 35.7         & 56.2        & 44.4          & 60.7          & 56.2         & 64.1         & 60.8                         \\ \hline
hu\_szeged    & 212            & 86.5        & 77.4        & 81.3         & 79.2         & 85.4        & 80.2          & 86.5          & 82.3         & 82.2         & 86.6                         \\ \hline
kk\_ktb       & 461            & 42.7        & 39.9        & 39.5         & 38.4         & 62.3        & 43.9          & 62.3          & 59.5         & 55.3         & 65.7                         \\ \hline
kmr\_mg       & 416            & 45.1        & 45.5        & 40.0         & 44.5         & 48.2        & 41.5          & 51.6          & 58.3         & 54.1         & 56.2                         \\ \hline
ko\_gsd       & 255            & 85.0        & 84.7        & 83.8         & 81.7         & 86.2        & 82.9          & 86.2          & 84.4         & 84.3         & 87.9                         \\ \hline
ko\_kaist     & 843            & 86.1        & 86.1        & 85.4         & 85.7         & 87.3        & 85.8          & 87.3          & 86.7         & 85.1         & 89.2                         \\ \hline
la\_ittb      & 133            & 85.1        & 86.0        & 85.0         & 83.3         & 86.2        & 83.9          & 86.2          & 85.5         & 84.6         & 87.0                         \\ \hline
la\_perseus   & 487            & 77.8        & 70.1        & 68.7         & 65.2         & 82.4        & 63.4          & 82.4          & 68.4         & 73.4         & 83.0                         \\ \hline
la\_proiel    & 114            & 86.2        & 53.2        & 51.9         & 78.4         & 84.8        & 58.8          & 86.2          & 84.2         & 80.0         & 84.9                         \\ \hline
lv\_lvtb      & 414            & 83.8        & 77.7        & 79.7         & 77.4         & 81.8        & 79.1          & 83.8          & 81.1         & 81.1         & 86.6                         \\ \hline
pl\_lfg       & 256            & 89.0        & 88.3        & 89.2         & 89.0         & 90.6        & 89.2          & 90.6          & 89.8         & 91.5         & 93.4                         \\ \hline
pl\_sz        & 235            & 88.8        & 87.2        & 86.9         & 86.8         & 90.1        & 87.0          & 90.1          & 88.4         & 89.0         & 91.7                         \\ \hline
pt\_bosque    & 94             & 89.9        & 85.8        & 84.9         & 85.5         & 91.2        & 82.8          & 91.2          & 90.1         & 87.7         & 92.4                         \\ \hline
sk\_snk       & 249            & 88.3        & 81.9        & 82.1         & 78.0         & 86.2        & 82.3          & 88.3          & 85.2         & 87.0         & 90.4                         \\ \hline
sl\_ssj       & 181            & 90.4        & 69.9        & 70.1         & 69.3         & 82.4        & 76.9          & 90.4          & 89.3         & 89.0         & 91.5                         \\ \hline
sme\_giella   & 507            & 69.2        & 70.7        & 69.9         & 69.9         & 71.2        & 61.6          & 71.2          & 71.3         & 74.4         & 72.6                         \\ \hline
sr\_set       & 121            & 88.3        & 88.0        & 87.9         & 86.2         & 90.5        & 87.7          & 90.5          & 88.8         & 90.1         & 93.2                         \\ \hline
sv\_lines     & 305            & 87.2        & 84.0        & 83.6         & 83.2         & 89.2        & 76.4          & 89.2          & 84.3         & 83.0         & 89.5                         \\ \hline
sv\_pud       & 433            & 84.6        & 83.3        & 83.4         & 77.4         & 90.6        & 82.1          & 90.6          & 84.3         & 85.2         & 91.3                         \\ \hline
sv\_talbanken & 387            & 88.2        & 85.0        & 85.5         & 82.9         & 91.6        & 85.3          & 91.6          & 85.9         & 86.6         & 91.9                         \\ \hline
tr\_imst      & 412            & 75.5        & 71.8        & 73.1         & 72.4         & 76.7        & 72.8          & 76.1          & 76.1         & 72.2         & 80.4                         \\ \hline
ug\_udt       & 346            & 82.2        & 78.7        & 78.5         & 78.2         & 84.3        & 73.3          & 82.2          & 81.3         & 77.0         & 82.2                         \\ \hline
uk\_iu        & 287            & 90.8        & 83.9        & 85.7         & 83.7         & 92.8        & 85.4          & 90.8          & 86.3         & 86.2         & 88.6                         \\ \hline
ur\_udtb      & 233            & 87.7        & 86.1        & 85.7         & 84.6         & 89.2        & 86.2          & 87.7          & 87.7         & 85.5         & 90.9                         \\ \hline
vi\_vtb       & 87             & 34.5        & 37.1        & 38.2         & 78.0         & 82.6        & 53.8          & 81.3          & 81.3         & 70.3         & 82.4                         \\ \hline
zh\_gsd       & 69             & 83.9        & 44.0        & 41.2         & 41.3         & 84.7        & 56.3          & 86.5          & 86.5         & 81.2         & 88.6                         \\ \hline
\end{tabular}
\label{tab:full_results3}
\end{table*}

\end{document}